\documentclass[10pt,twocolumn,letterpaper]{article}

\usepackage{cvpr}              

\usepackage{graphicx}
\usepackage{amsmath}
\usepackage{amssymb}
\usepackage{booktabs}

\usepackage[accsupp]{axessibility}

\usepackage{algorithm}
\usepackage{algorithmic}

\usepackage{multirow}
\usepackage[pagebackref,breaklinks,colorlinks]{hyperref}

\usepackage[capitalize]{cleveref}
\crefname{section}{Sec.}{Secs.}
\Crefname{section}{Section}{Sections}
\Crefname{table}{Table}{Tables}
\crefname{table}{Tab.}{Tabs.}

\def\cvprPaperID{7788} 
\def\confName{CVPR}
\def\confYear{2023}

\begin{document}

\title{Global and Local Mixture Consistency Cumulative Learning for Long-tailed Visual Recognitions}

\author{
	Fei Du\textsuperscript{1,2,3}, Peng Yang\textsuperscript{1,3}, Qi Jia\textsuperscript{1,3}, Fengtao Nan\textsuperscript{1,2,3}, Xiaoting Chen\textsuperscript{1,3}, Yun Yang\textsuperscript{1,3}\thanks{ Corresponding author} \\
	\textsuperscript{1}National Pilot School of Software, Yunnan University, Kunming, China\\
	\textsuperscript{2}School of Information Science and Engineering, Yunnan University, Kunming, China\\
	\textsuperscript{3}Yunnan Key Laboratory of Software Engineering\\
	{\tt\small \{dufei,yangpeng,jiaqi,fengtaonan,chenxiaoting\}@mail.ynu.edu.cn yangyun@ynu.edu.cn}}
\maketitle

\begin{abstract}
In this paper, our goal is to design a simple learning paradigm for long-tail visual recognition, which not only improves the robustness of the feature extractor but also alleviates the bias of the classifier towards head classes while reducing the training skills and overhead. We propose an efficient one-stage training strategy for long-tailed visual recognition called Global and Local Mixture Consistency cumulative learning (GLMC). Our core ideas are twofold: (1) a global and local mixture consistency loss improves the robustness of the feature extractor. Specifically, we generate two augmented batches by the global MixUp and local CutMix from the same batch data, respectively, and then use cosine similarity to minimize the difference. (2) A cumulative head-tail soft label reweighted loss mitigates the head class bias problem. We use empirical class frequencies to reweight the mixed label of the head-tail class for long-tailed data and then balance the conventional loss and the rebalanced loss with a coefficient accumulated by epochs. Our approach achieves state-of-the-art accuracy on CIFAR10-LT, CIFAR100-LT, and ImageNet-LT datasets. Additional experiments on balanced ImageNet and CIFAR demonstrate that GLMC can significantly improve the generalization of backbones.  Code is made publicly available at \href{https://github.com/ynu-yangpeng/GLMC}{https://github.com/ynu-yangpeng/GLMC}.
\end{abstract}

\section{Introduction}
\label{sec:intro}

Thanks to the available large-scale datasets, \eg, ImageNet \cite{deng2009imagenet}, MS COCO \cite{lin2014microsoft}, and Places \cite{zhou2017places} Database, deep neural networks have achieved dominant results in image recognition \cite{he2016deep}. Distinct from these well-designed balanced datasets, data naturally follows long-tail distribution in real-world scenarios, where a small number of head classes occupy most of the samples. In contrast, dominant tail classes only have a few samples. Moreover, the tail classes are critical for some applications, such as medical diagnosis and autonomous driving. Unfortunately, learning directly from long-tailed data may cause model predictions to over-bias toward the head classes.

\begin{figure}[]
	\centering
	\includegraphics[width=0.5\textwidth]{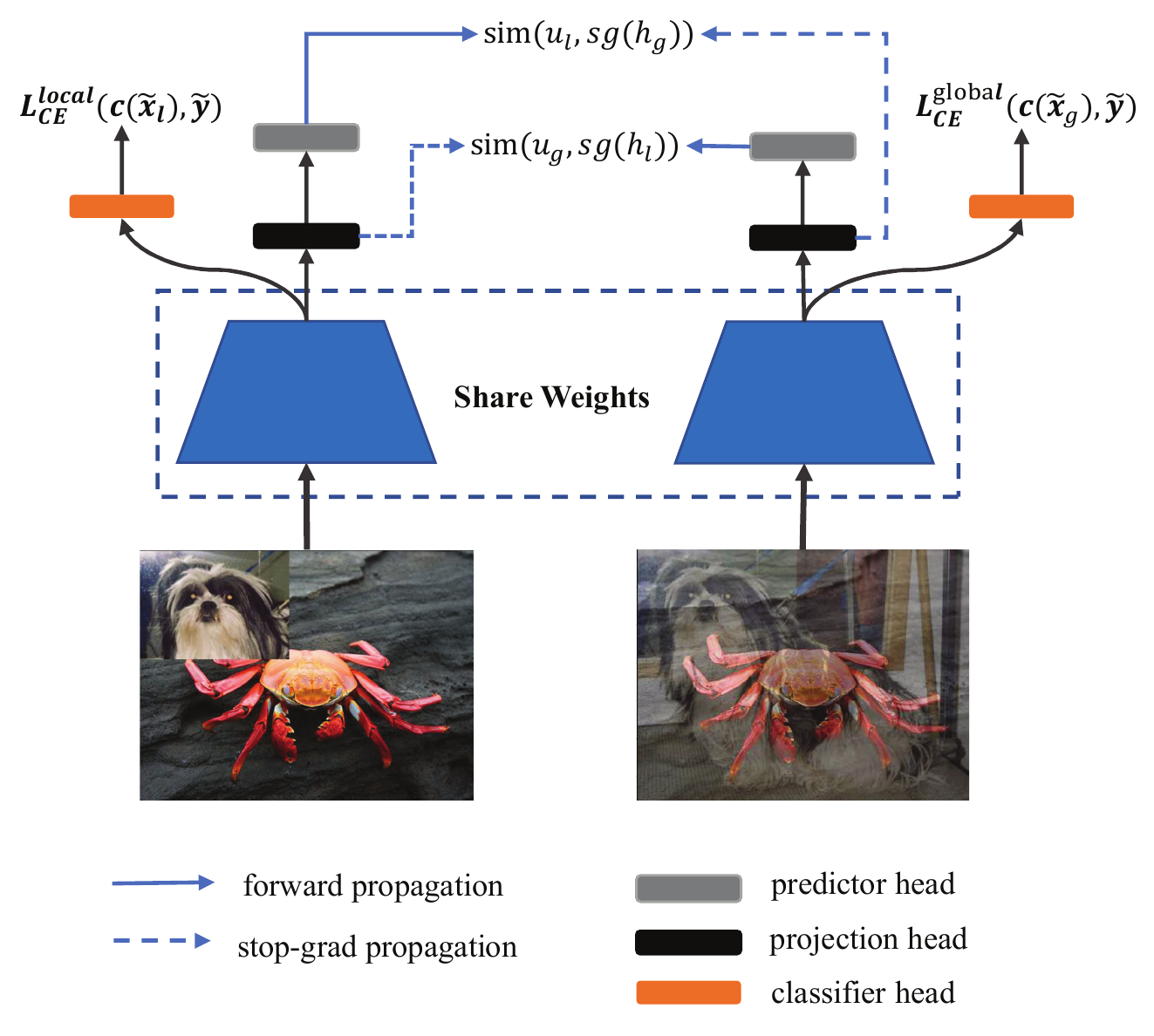}
	\caption{An overview of our GLMC: two types of mixed-label augmented images are processed by an encoder network and a projection head to obtain the representation $h_g$ and $h_l$. Then a prediction head transforms the two representations to output $u_g$ and $u_l$. We minimize their negative cosine similarity as an auxiliary loss in the supervised loss. $sg(\cdot)$ denotes stop gradient operation.
}
\label{fig:GLMC}    
\end{figure}

There are two classical rebalanced strategies for long-tailed distribution, including resampling training data \cite{han2005borderline,chu2020feature,shen2016relay} and designing cost-sensitive reweighting loss functions \cite{khan2017cost,cao2019learning}. For the resampling methods, the core idea is to oversample the tail class data or undersample the head classes in the SGD mini-batch to balance training. As for the reweighting strategy, it mainly increases the loss weight of the tail classes to strengthen the tail class. However, learning to rebalance the tail classes directly would damage the original distribution \cite{zhou2020bbn} of the long-tailed data, either increasing the risk of overfitting in the tail classes or sacrificing the performance of the head classes. Therefore, these methods usually adopt a two-stage training process \cite{zhou2020bbn,cao2019learning,alshammari2022long} to decouple the representation learning and classifier fine-tuning: the first stage trains the feature extractor on the original data distribution, then fixes the representation and trains a balanced classifier. Although multi-stage training significantly improves the performance of long-tail recognition, it also negatively increases the training tricks and overhead.

In this paper, our goal is to design a simple learning paradigm for long-tail visual recognition, which not only improves the robustness of the feature extractor but also alleviates the bias of the classifier towards head classes while reducing the training skills and overhead. For improving representation robustness, recent contrastive learning techniques \cite{kang2020exploring,zhu2022balanced,cui2021parametric,li2021self} that learn the consistency of augmented data pairs have achieved excellence. Still, they typically train the network in a two-stage manner, which does not meet our simplification goals, so we modify them as an auxiliary loss in our supervision loss. For head class bias problems, the typical approach is to initialize a new classifier for resampling or reweighting training. Inspired by the cumulative weighted rebalancing \cite{zhou2020bbn} branch strategy, we adopt a more efficient adaptive method to balance the conventional and reweighted classification loss.

Based on the above analysis, we propose an efficient one-stage training strategy for long-tailed visual recognition called Global and Local Mixture Consistency cumulative learning (GLMC). Our core ideas are twofold: (1) a global and local mixture consistency loss improves the robustness of the model. Specifically, we generate two augmented batches by the global MixUp and local CutMix from the same batch data, respectively, and then use cosine similarity to minimize the difference. (2) A cumulative head-tail soft label reweighted loss mitigates the head class bias problem. Specifically, we use empirical class frequencies to reweight the mixed label of the head-tail class for long-tailed data and then balance the conventional loss and the rebalanced loss with a coefficient accumulated by epochs.

Our method is mainly evaluated in three widely used long-tail image classification benchmark datasets, which include CIFAR10-LT, CIFAR100-LT, and ImageNet-LT datasets. Extensive experiments show that our approach outperforms other methods by a large margin, which verifies the effectiveness of our proposed training scheme. Additional experiments on balanced ImageNet and CIFAR demonstrate that GLMC can significantly improve the generalization of backbones. The main contributions of our work can be summarized as follows:

\begin{itemize}
	
	\item We propose an efficient one-stage training strategy called Global and Local Mixture Consistency cumulative learning framework (GLMC), which can effectively improve the generalization of the backbone for long-tailed visual recognition.
	
	\item GLMC does not require negative sample pairs or large batches and can be as an auxiliary loss added in supervised loss.

	\item Our GLMC achieves state-of-the-art performance on three challenging long-tailed recognition benchmarks, including CIFAR10-LT, CIFAR100-LT, and ImageNet-LT datasets. Moreover, experimental results on full ImageNet and CIFAR validate the effectiveness of GLMC under a balanced setting.
	
\end{itemize}

\section{Related Work}

\subsection{Contrastive Representation Learning}
The recent renaissance of self-supervised learning is expected to obtain a general and transferrable feature representation by learning pretext tasks.  For computer vision, these pretext tasks include rotation prediction \cite{komodakis2018unsupervised}, relative position prediction of image patches \cite{doersch2015unsupervised}, solving jigsaw puzzles \cite{noroozi2016unsupervised}, and image colorization\cite{zhang2016colorful,larsson2017colorization}. However, these pretext tasks are usually domain-specific, which limits the generality of learned representations.  

Contrastive learning is a significant branch of self-supervised learning. Its pretext task is to bring two augmented images (seen as positive samples) of one image closer than the negative samples in the representation space. Recent works \cite{oord2018representation,hjelm2018learning,tian2020contrastive} have attempted to learn the embedding of images by maximizing the mutual information of two views of an image between latent representations. However, their success relies on a large number of negative samples. To handle this issue, BYOL \cite{grill2020bootstrap} removes the negative samples and directly predicts the output of one view from another with a momentum encoder to avoid collapsing. Instead of using a momentum encoder, Simsiam \cite{chen2021exploring} adopts siamese networks to maximize the cosine similarity between two augmentations of one image with a simple stop-gradient technique to avoid collapsing. 

For long-tail recognition, there have been numerous works \cite{kang2020exploring,zhu2022balanced,cui2021parametric,li2021self} to obtain a balanced representation space by introducing a contrastive loss. However, they usually require a multi-stage pipeline and large batches of negative examples for training, which negatively increases training skills and overhead. Our method learns the consistency of the mixed image by cosine similarity, and this method is conveniently added to the supervised training in an auxiliary loss way. Moreover, our approach neither uses negative pairs nor a momentum encoder and does not rely on large-batch training.

\subsection{Class Rebalance learning}

Rebalance training has been widely studied in long-tail recognition. Its core idea is to strengthen the tail class by oversampling \cite{chawla2002smote,han2005borderline} or increasing weight\cite{byrd2019effect,cui2019class,zhang2021distribution}. However, over-learning the tail class will also increase the risk of overfitting \cite{zhou2020bbn}. Conversely, under-sampling or reducing weight in the head class will sacrifice the performance of head classes. Recent studies \cite{kang2019decoupling,zhou2020bbn} have shown that directly training the rebalancing strategy would degrade the performance of representation extraction, so some multi-stage training methods \cite{kang2019decoupling,zhou2020bbn,alshammari2022long} decouple the training of representation learning and classifier for long-tail recognition. For representation learning, self-supervised-based \cite{kang2020exploring,zhu2022balanced,li2021self} and augmentation-based \cite{chen2022imagine,park2022majority} methods can improve robustness to long-tailed distributions. And for the rebalanced classifier, such as multi-experts\cite{wang2020long,li2022trustworthy}, reweighted classifiers\cite{alshammari2022long}, and label-distribution-aware \cite{cao2019learning}, all can effectively enhance the performance of tail classes. Further, \cite{zhou2020bbn} proposed a unified Bilateral-Branch Network (BBN) that adaptively adjusts the conventional learning branch and the reversed sampling branch through a cumulative learning strategy. Moreover, we follow BBN to weight the mixed labels of long-tailed data adaptively and do not require an ensemble during testing.

\section{The Proposed Method}
In this section, we provide a detailed description of our GLMC framework. First, we present an overview of our framework in Sec.\ref{framework}. Then, we introduce how to learn global and local mixture consistency by maximizing the cosine similarity of two mixed images in Sec.\ref{Mixture Consistency learning}. Next, we propose a cumulative class-balanced strategy to weight long-tailed data labels progressively in Sec.\ref{class-balanced learning}. Finally,
we introduce how to optionally use MaxNorm \cite{hinton2012improving,alshammari2022long} to finetune the classifier weights in Sec.\ref{finetune w}.
\begin{figure*}[htbp]
	\centering
	\includegraphics[width=0.9\textwidth]{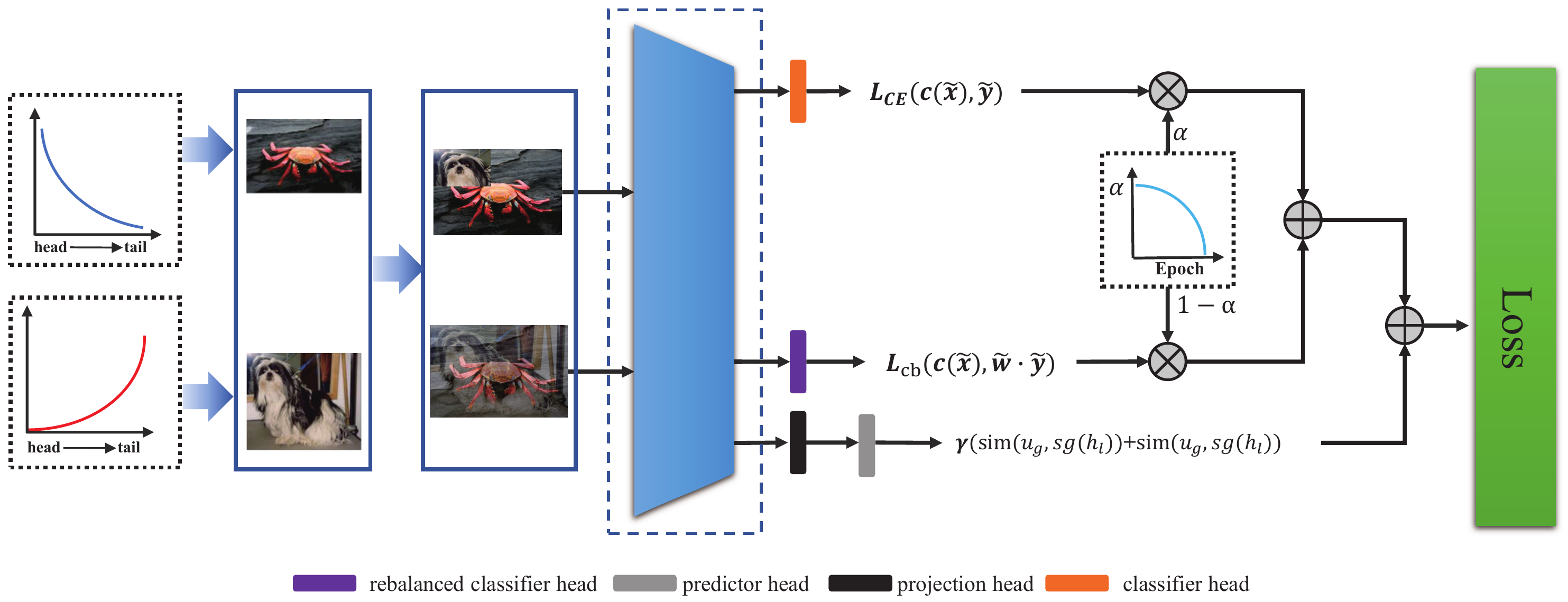}
	\caption{An illustration of the cumulative class-balanced learning pipeline. We apply uniform and reversed samplers to obtain head and tail data, and then they are synthesized into head-tail mixture samples by MixUp and CutMix. The cumulative learning strategy adaptively weights the rebalanced classifier and the conventional classifier by epochs.}
	\label{fig:cumulative learning strategy}    
\end{figure*}

\subsection{Overall Framework}
\label{framework}
Our framework is divided into the following six major components: 
\begin{itemize}
	\item A stochastic mixed-label data augmentation module $Aug(x,y)$. For each input batch samples, $Aug(x,y)$ transforms $x$ and their labels $y$ in global and local augmentations pairs, respectively.
	\item An encoder (e.g., ResNet) $f(\tilde{x})$ that extracts representation vectors $r$ from the augmented samples $\tilde{x}$.
	\item A projection $proj(x)$ that maps vectors $r$ to lower dimension representations $h$. The projection is simply a fully connected layer. Its output has no activation function. 
	\item A predictor $pred(x)$ that maps the output of projection to the contrastive space. The predictor also a fully connected layer and has no activation function. 
	\item A linear conventional classifier head $c(x)$ that maps vectors $r$ to category space. The classifier head calculates mixed cross entropy loss with the original data distribution.
	\item (optional) A linear rebalanced classifier head $cb(x)$ that maps vectors $r$ to rebalanced category space. The rebalanced classifier calculates mixed cross entropy loss with the reweighted data distribution.
\end{itemize}
Note that only the rebalanced classifier $cb(x)$ is retained at the end of training for the long-tailed recognition, while the predictor, projection, and conventional classifier head will be removed. However, for the balanced dataset, the rebalanced classifier $cb(x)$ is not needed.
\subsection{Global and Local Mixture Consistency Learning}
\label{Mixture Consistency learning}

In supervised deep learning, the model is usually divided into two parts: an encoder and a linear classifier. And the classifiers are label-biased and rely heavily on the quality of representations. Therefore, improving the generalization ability of the encoder will significantly improve the final classification accuracy of the long-tailed challenge. Inspired by self-supervised learning to improve representation by learning additional pretext tasks, as illustrated in Fig.\ref{fig:GLMC}, we train the encoder using a standard supervised task and a self-supervised task in a multi-task learning way. Further, unlike simple pretext tasks such as rotation prediction, image colorization, etc., following the global and local ideas\cite{7112527}, we expect to learn the global-local consistency through the strong data augmentation method MixUp \cite{zhang2017mixup} and CutMix \cite{yun2019cutmix}.

\noindent \textbf{Global Mixture.} MixUp is a global mixed-label data augmentation method that generates mixture samples by mixing two images of different classes. For a pair of two images and their labels probabilities $(x_i,p_i)$ and $(x_j,p_j)$, we calculate $(\tilde{x}_g,\tilde{p}_g)$ by
\begin{equation}
\label{eq:mixup}
\begin{aligned}
\lambda &\sim Beta(\beta ,\beta), \\
\tilde{x}_g &= \lambda x_i + (1 -\lambda)x_j, \\
\tilde{p}_g &= \lambda p_i + (1 -\lambda)p_j. 
\end{aligned}
\end{equation}
where $\lambda$ is sampled from a $Beta$ distribution parameterized by the $\beta$ hyper-parameter. Note that $p$ are one-hot vectors.

\noindent \textbf{Local Mixture.} Different from MixUp, CutMix combines two images by locally replacing the image region with a patch from another training image. We define the combining operation as
\begin{equation}
\label{eq:cutmix}
\begin{aligned}
\tilde{x}_l &= \boldsymbol{M}\odot x_i + (\boldsymbol{1} -\boldsymbol{M})\odot x_j.
\end{aligned}
\end{equation}
where $\boldsymbol{M}\in \{0,1\}^{W\times H}$ denotes the randomly selected pixel patch from the image $x_i$ and pasted on $x_j$, $\boldsymbol{1}$ is a binary mask filled with ones, and $\odot$ is element-wise multiplication. Concretely, we sample the bounding box coordinates $B=(r_x,r_y,r_w,r_h)$ indicating the cropping regions on $x_i$ and $x_j$. The box coordinates are uniformly sampled according to
\begin{equation}
\label{eq:6}
\begin{aligned}
r_x&\sim Uniform(0,W), r_w=W\sqrt{1-\lambda} \\
r_y&\sim Uniform(0,H), r_h=H\sqrt{1-\lambda} 
\end{aligned}
\end{equation}
where $\lambda$ is also sampled from the $Beta(\beta,\beta)$, and their mixed labels are the same as MixUp.

\textbf{Self-Supervised Learning Branch.} Previous works require large batches of negative samples \cite{hjelm2018learning,tian2020contrastive} or a memory bank \cite{he2020momentum} to train the network. That makes it difficult to apply to devices with limited memory. For simplicity, our goal is to maximize the cosine similarity of global and local mixtures in representation space to obtain contrastive consistency. Specifically, the two types of augmented images are processed by an encoder network and a projection head to obtain the representation $h_g$ and $h_l$. Then a prediction head transforms the two representations to output $u_g$ and $u_l$. We minimize their negative cosine similarity:
\begin{equation}
\label{eq:9}
sim(u_g,h_l)=-\frac{u_g}{\left \| u_g \right \|}\cdot \frac{h_l}{\left \| h_l \right \|}
\end{equation}
where $\left \| \cdot \right \|$ is  $l_2$ normalization. An undesired trivial solution to minimize the negative cosine similarity of augmented images is all outputs “collapsing” to a constant. Following SimSiam \cite{chen2021exploring}, we use a stop gradient operation to prevent collapsing. The SimSiam loss function is defined as:
\begin{equation}
\label{eq:10}
\pounds_{sim}=sim(u_g,sg(h_l))+sim(u_l,sg(h_g))
\end{equation}
this means that $h_l$ and $h_g$ are treated as a constant.

\textbf{Supervised Learning Branch.} After constructing the global and local augmented data pair $(\tilde{x}_g;\tilde{p}_g)$ and $(\tilde{x}_l;\tilde{p}_l)$, we calculate the mixed-label cross-entropy loss:
\begin{equation}
\label{eq:7}
\begin{aligned}
\pounds_{c} =-\frac{1}{2N}\sum_{i=1}^{N}(\tilde{p}_{g}^{i}(logf(\tilde{x}_{g}^{i}))+\tilde{p}_{l}^{i}(logf(\tilde{x}_{l}^{i})))
\end{aligned}
\end{equation}
where $N$ denote the sampling batch size and $f(\cdot)$ denote predicted probability of $\tilde{x}$. Note that a batch of images is augmented into a global and local mixture so that the actual batch size will be twice the sampling size.

\begin{algorithm}[tb]
	
	\caption{Learning algorithm of our proposed GLMC}
	\label{alg:algorithm}
	\textbf{Input}: Training Dataset $D=\{(x_i,y_i,w_i)\}_{i=1}^N$\\
	\textbf{Parameter}: $Encoder(\cdot)$ denotes feature extractor; $proj(\cdot)$ and $pred(\cdot)$ denote projection and predictor; $c(\cdot)$ and $cb(\cdot)$ denote convention classifier and rebalanced classifier; $T_{max}$ is the Maximum training epoch; $sg(\cdot)$ denotes stop gradient operation.
	\begin{algorithmic}[1] 
		\STATE \textbf{for} $T=1$ in $T_{max}$ \textbf{do} 
		\STATE \quad $\alpha \leftarrow 1 - (\frac{T}{T_{max}})^2$ 
		\STATE \quad \textbf{for} $(x,y,w)$ in $D$ \textbf{do} 
		
		\STATE \quad\quad $\lambda \leftarrow Beta(\beta ,\beta)$ 
		\STATE \quad\quad $(\tilde{x}_g,\tilde{p}_g,\tilde{w}_g) \leftarrow MixUp(x,y,w,\lambda) $
		\STATE \quad\quad $(\tilde{x}_l,\tilde{p}_l,\tilde{w}_l) \leftarrow CutMix(x,y,w,\lambda) $  \\// Generate global and local mixed augmented data.
		\STATE \quad\quad $r_g,r_l \leftarrow Encoder(\tilde{x}_g), Encoder(\tilde{x}_l) $ 
		\STATE \quad\quad $h_g,h_l \leftarrow proj(r_g), proj(r_l) $   \\// Map representation $r_g$ and $r_l$ to vector  $h_g$ and $h_l$.
		\STATE \quad\quad $u_g,u_l \leftarrow pred(h_g), pred(h_l) $ \\// Map representation $h_g$ and  $h_l$ to contrastive space  $u_g$ and $u_l$.
		\STATE \quad\quad $\pounds_{sim} \leftarrow sim(u_g,sg(h_l)) + sim(u_l,sg(h_g))  $ \\// Calculate global and local mixture similarity.
		
		\STATE \quad\quad $ p^c_g,p^c_l \leftarrow Sofmatx(c(r_g)),Sofmatx(c(r_l)) $\\// Calculate the classification probability of the convention branch
		\STATE \quad\quad $\pounds_{c}\leftarrow  \pounds(\tilde{p},p^c_g)+\pounds(\tilde{p},p^c_l)$\\// Calculate the classification loss

		\STATE \quad\quad $ p^{cb}_g,p^{cb}_l \leftarrow Sofmatx(cb(r_g)),Sofmatx(cb(r_l)) $ 
		\\// Calculate the classification probability of the rebalance branch
		\STATE \quad\quad $\pounds_{cb}\leftarrow  \pounds(\tilde{p},p^{cb}_g)+\pounds(\tilde{p},p^{cb}_l)$ \\// Calculate the rebalanced classification loss 
		\STATE \quad\quad $ \pounds_{total}= \alpha \pounds_{c}+  (1-\alpha)\pounds_{cb} + \gamma\pounds_{sim}$	\\// Calculate the total loss 
		\STATE \quad\quad  Update model parameters by minimizing $\pounds_{total}$ 
		\STATE \quad\textbf{end for}
		\STATE \textbf{end for}
	\end{algorithmic}
\end{algorithm}

\begin{table*}[ht]
	\centering
	\caption{ Top-1 accuracy (\%) of ResNet-32 on CIFAR-10-LT and CIFAR-100-LT with different imbalance factors [100, 50, 10]. GLMC consistently outperformed the previous best method only in the one-stage.}
	\label{table_cifar}
	\begin{tabular}{c|l|lll|lll} 
		\hline
		& \multirow{2}{*}{Method } & \multicolumn{3}{c|}{CIFAR-10-LT} & \multicolumn{3}{c}{CIFAR-100-LT}                  \\ 
		\cline{3-8}
		&                          & IF=100 & 50    & 10              & 100            & 50             & 10              \\ 
		\hline
		& CE                       & 70.4   & 74.8  & 86.4            & 38.3           & 43.9           & 55.7            \\ 
		\hline
		\multirow{4}{*}{rebalance classifier}         & BBN \cite{zhou2020bbn}         & 79.82          & 82.18 & 88.32           & 42.56          & 47.02         & 59.12     \\
		& CB-Focal   \cite{cui2019class}                 & 74.6       & 79.3     & 87.1            & 39.6           & 45.2           & 58              \\
		& LogitAjust  \cite{menon2020long}               & 80.92      & -        & -               & 42.01          & 47.03          & 57.74           \\
		& weight~balancing \cite{alshammari2022long}  & -       & -        & -               & 53.35          & 57.71          & 68.67           \\ 
		\hline
		\multirow{3}{*}{augmentation}                 & Mixup \cite{zhang2017mixup}                & 73.06  & 77.82 & 87.1            & 39.54       & 54.99        & 58.02    \\
		& RISDA  \cite{chen2022imagine}               & 79.89         & 79.89    & 79.89           & 50.16          & 53.84          & 62.38           \\
		& CMO  \cite{park2022majority}                & -      & -    & -        & 47.2            & 51.7           & 58.4            \\ 
		\hline
		\multirow{5}{*}{self-supervised pretraining}  & KCL  \cite{kang2020exploring}                      & 77.6   & 81.7  & 88      & 42.8      & 46.3        & 57.6      \\
		& TSC \cite{li2022targeted}                   & 79.7   & 82.9  & 88.7            & 42.8           & 46.3           & 57.6           \\
		& BCL  \cite{zhu2022balanced}                 & 84.32  & 87.24 & 91.12           & 51.93          & 56.59          & 64.87       \\
		& PaCo \cite{cui2021parametric}               & -      & -     & -               & 52             & 56             & 64.2       \\
		& SSD   \cite{li2021self}                     & -      & -     & -               & 46             & 50.5           & 62.3              \\ 
		\hline
		\multirow{2}{*}{ensemble classifier}                     & RIDE~(3~experts)~+~CMO \cite{park2022majority}  & -      & -     & -        & 50       & 53         & 60.2     \\
		& RIDE~(3~experts)  \cite{wang2020long}        & -      & -     & -               & 48.6           & 51.4           & 59.8       \\ 
		\hline
	one-stage training	& \textbf{ours}            &  \textbf{87.75}      &   \textbf{90.18}    &    \textbf{94.04}            & \textbf{55.88} & \textbf{61.08} & \textbf{70.74}  \\ 
		\hline
	finetune  classifier	& \textbf{ours~+~MaxNorm\cite{alshammari2022long} }       &   \textbf{87.57}     &  \textbf{90.22}     &     \textbf{94.03}            & \textbf{57.11} & \textbf{62.32} & \textbf{72.33}  \\
		\hline
	\end{tabular}

\end{table*}

\subsection{Cumulative Class-Balanced Learning}
\label{class-balanced learning}
\textbf{Class-Balanced learning.} The design principle of class reweighting is to introduce a weighting factor inversely proportional to the label frequency and then strengthen the learning of the minority class. Following \cite{zhang2021distribution}, the weighting factor $w_i$ is define as: 
\begin{equation}
\label{eq:12}
w_i= \frac{C\cdot(1/r_i)^k}{\sum_{i=1}^{C}(1/r_i)^k}
\end{equation}
where $r_i$ is the i-th class frequencies of the training dataset, and $k$ is a hyper-parameter to scale the gap between the head and tail classes. Note that $k=0$ corresponds to no re-weighting and $k=1$ corresponds to class-balanced method \cite{cui2019class}. We change the scalar weights to the one-hot vectors form and mix the weight vectors of the two images:
\begin{equation}
\label{eq:17}
\tilde{w} = \lambda w_i + (1 -\lambda)w_j.
\end{equation}
Formally, given a train dataset $D=\{(x_i,y_i,w_i)\}_{i=1}^N$, the rebalanced loss can be written as:
\begin{equation}
\label{eq:13}
\pounds_{cb} =-\frac{1}{2N}\sum_{i=1}^{N}\tilde{w}^i(\tilde{p}_{g}^{i}(logf(\tilde{x}_{g}^{i}))+\tilde{p}_{l}^{i}(logf(\tilde{x}_{l}^{i})))
\end{equation}
where $f(\tilde{x})$ and $\tilde{w}$ denote predicted probability and weighting factor of mixed image $\tilde{x}$, respectively. Note that the global and local mixed image have the same mixed weights.

\textbf{Cumulative Class-Balanced Learning.} As illustrated in Fig.\ref{fig:cumulative learning strategy}, we use the bilateral branches structure to learn the rebalance branch adaptively. But unlike BBN \cite{zhou2020bbn}, our cumulative learning strategy is imposed on the loss function instead of the fully connected layer weights and uses reweighting instead of resampling for learning. Concretely, the loss $\pounds_{c}$ of the unweighted classification branch is multiplied by $\alpha$, and the rebalanced loss $ \pounds_{cb}$ is multiplied by $1-\alpha$. $\alpha$ automatically decreases as the current training epochs $T$ increase:
\begin{equation}
\label{eq:15}
\alpha = 1 - (\frac{T}{T_{max}})^2
\end{equation}
where $T_{max}$ is the maximum training epoch.

Finally, the total loss is defined as a combination of loss $L_{sup}$, $L_{cb}$, and $L_{sim}$:
\begin{equation}
\label{eq:16}
\pounds_{total}= \alpha \pounds_{c}+  (1-\alpha)\pounds_{cb} + \gamma \pounds_{sim}
\end{equation}
where $\gamma$ is a hyperparameter that controls  $L_{sim}$ loss. The default value is 10.

\subsection{Finetuning Classifier Weights}
\label{finetune w}
\cite{alshammari2022long} investigate that the classifier weights would be heavily biased toward the head classes when faced with long-tail data. Therefore, we optionally use MaxNorm \cite{hinton2012improving,alshammari2022long} to finetune the classifier in the second stage. Specifically, MaxNorm restricts  weight norms within a ball of radius $\delta$:
\begin{equation}
\label{eq:19}
\Theta^\ast  = arg\underset{\Theta}{min}F(\Theta;D),\; s.t.||\theta _k||^2_2\leq \delta^2
\end{equation}
this can be solved by applying projected gradient descent (PGD). For each epoch (or iteration), PGD first computes an updated $\theta_k $ and then projects it onto the norm ball:
\begin{equation}
\theta_k\leftarrow min(1,\delta/||\theta_k||_2)*\theta_k
\end{equation}

\section{Experiments}
In this section, we evaluate the proposed GLMC on three widely used long-tailed benchmarks: CIFAR-10-LT, CIFAR-100-LT, and ImageNet-LT. We also conduct a series of ablation studies to assess each component of GLMC's importance fully.

\begin{table*}[]
	\centering
		\caption{Top-1 accuracy (\%) on ImageNet-LT dataset. Comparison to the state-of-the-art methods with different backbone. $\dagger$ denotes results reproduced by \cite{zhu2022balanced} with 180 epochs.}
	\label{tabel-imageNetLT}
	\begin{tabular}{l|c|cccc}
		\hline
		\multirow{2}{*}{Method} & \multirow{2}{*}{Backbone} & \multicolumn{4}{c}{ImageNet-LT} \\ \cline{3-6} 
		&                           & Many   & Med   & Few   & All    \\ \hline
		CE                      & ResNet-50                 & 64     & 33.8  & 5.8   & 41.6   \\
		CB-Focal \cite{cui2019class}        & ResNet-50                 & 39.6   & 32.7  & 16.8  & 33.2   \\
		LDAM  \cite{cao2019learning}        & ResNet-50                 & 60.4   & 46.9  & 30.7  & 49.8   \\
		KCL   \cite{kang2020exploring}      & ResNet-50                 & 61.8   & 49.4  & 30.9  & 51.5   \\
		TSC   \cite{li2022targeted}         & ResNet-50                 & 63.5   & 49.7  & 30.4  & 52.4   \\
		RISDA \cite{chen2022imagine}        & ResNet-50                 &  -     &  -    &  -    & 49.3  \\
		BCL (90 epochs) \cite{zhu2022balanced}         & ResNeXt-50                & 67.2   & 53.9  & 36.5  & 56.7   \\
		BCL (180 epochs)  \cite{zhu2022balanced}         & ResNeXt-50                & 67.9   & 54.2  & 36.6  & 57.1   \\
		PaCo$^\dagger$ (180 epochs)  \cite{cui2021parametric}       & ResNeXt-50                 & 64.4   & 55.7  & 33.7  & 56.0   \\
		Balanced Softmax$^\dagger$ (180 epochs)  \cite{ren2020balanced}       & ResNeXt-50                 & 65.8   & 53.2  & 34.1  & 55.4   \\
		SSD  \cite{li2021self}              & ResNeXt-50                & 66.8   & 53.1  & 35.4  & 56     \\
		RIDE~(3~experts)~+~CMO \cite{park2022majority} & ResNet-50                 & 66.4   & 53.9  & 35.6  & 56.2   \\
		RIDE (3 experts) \cite{wang2020long}        & Swin-S                    & 66.9   & 52.8  & 37.4  & 56     \\
		weight~balancing~+~MaxNorm \cite{alshammari2022long}  & ResNeXt-50                & 62.5   & 50.4  & 41.5  & 53.9   \\ \hline
		ours                    &  \multirow{3}{*}{ResNeXt-50 }                      & \textbf{70.1}     & 52.4    & 30.4    & 56.3     \\
		ours~+~MaxNorm \cite{alshammari2022long}          &                         & 60.8     & \textbf{55.9}    & \textbf{45.5}    & 56.7   \\ 
		ours~+~BS  \cite{ren2020balanced}         &                        & 64.76     & 55.67    & 42.19    & \textbf{57.21}   \\ \hline
	\end{tabular}

\end{table*}
\subsection{Experiment setup}
\textbf{Datasets.} Following \cite{yue2016deep}, we modify the balanced CIFAR10, CIFAR100, and ImageNet-2012 dataset to the uneven setting (named CIFAR10-LT, CIFAR100-LT, and ImageNet-LT) by utilizing the exponential decay function $n = n_i\mu^i$, where $i$ is the class index (0-indexed), $n_i$ is the original number of training images and $\mu \in (0,1)$. The imbalanced factor $\beta$ is defined by $\beta = N_{max}/N_{min}$, which reflects the degree of imbalance in the data. CIFAR10-LT and CIFAR100-LT are divided into three types of train datasets, and each dataset has a different imbalance factor [100,50,10]. ImageNet-LT has a 256 imbalance factor. The most frequent class includes 1280 samples, while the least contains only 5.

\textbf{Network architectures.} For a fair comparison with recent works, we follow \cite{alshammari2022long,zhu2022balanced,cui2021parametric} to use ResNet-32 \cite{he2016deep} on CIFAR10-LT and CIFAR100-LT, ResNet-50 \cite{he2016deep} and ResNeXt-50-32x4d \cite{xie2017aggregated} on ImageNet-LT. The main ablation experiment was performed using ResNet-32 on the CIFAR100 dataset.

\textbf{Evaluation protocol.} For each dataset, we train them on the imbalanced training set and evaluate them in the balanced validation/test set. Following \cite{liu2019large,alshammari2022long}, we further report accuracy on three splits of classes, Many-shot classes (training samples $\textgreater$ 100), Medium-shot (training samples 20 $\sim$ 100) and Few-shot (training samples $\leq$ 20), to comprehensively evaluate our model.

\textbf{Implementation.} We train our models using the PyTorch toolbox \cite{paszke2017automatic} on GeForce RTX 3090 GPUs. All models are implemented by the SGD optimizer with a momentum of 0.9 and gradually decay learning rate with a cosine annealing scheduler, and the batch size is 128. For CIFAR10-LT and CIFAR100-LT, the initial learning rate is 0.01, and the weight decay rate is 5e-3. For ImageNet-LT, the initial learning rate is 0.1, and the weight decay rate is 2e-4. We also use random horizontal flipping and cropping as simple augmentation.

\begin{table}
	\centering
	\caption{Top-1 accuracy (\%) on full ImageNet dataset with ResNet-50 backbone.}
	\label{table-full-img}
	\begin{tabular}{l|l|l} 
		\hline
		Method   & Augmentation   & Top-1 acc  \\ 
		\hline
		vanilla & Simple Augment & 76.4       \\ 
		
		vanilla & MixUp \cite{zhang2017mixup}         & 77.9       \\
		vanilla & CutMix \cite{yun2019cutmix}        & 78.6       \\ 
		\hline
		Supcon \cite{khosla2020supervised}  & RandAugment    & 78.4       \\
		PaCo \cite{cui2021parametric}    & Simple Augment & 78.7       \\
		PaCo \cite{cui2021parametric}    & RandAugment    & 79.3       \\ 
		\hline
		ours     & MixUp~+~CutMix &     \textbf{80.2}       \\ 
		\hline
	\end{tabular}
	
\end{table}

\subsection{Long-tailed Benchmark Results}
\textbf{Compared Methods.} Since the field of LTR is developing rapidly and has many branches, we choose recently published representative methods of different types for comparison. For example, SSD \cite{li2021self}, PaCo \cite{cui2021parametric}, KCL \cite{kang2020exploring}, BCL \cite{zhu2022balanced}, and TSC \cite{li2022targeted} use contrastive learning or self-supervised methods to train balanced representations. RIDE \cite{wang2020long} combines multiple experts for prediction; RISDA \cite{chen2022imagine} and CMO \cite{park2022majority} apply strong data augmentation techniques to improve robustness;  Weight Balancing \cite{alshammari2022long} is a typical two-stage training method.

\noindent\textbf{Results on CIFAR10-LT and CIFAR100-LT.} We conduct extensive experiments to compare GLMC with state-of-the-art baselines on long-tailed CIFAR10 and CIFAR100 datasets by setting three imbalanced ratios: 10, 50, and 100. Table \ref{table_cifar} reports the Top-1 accuracy of various methods on CIFAR-10-LT and CIFAR-100-LT. We can see that our GLMC consistently achieves the best results on all datasets, whether using one-stage training or a two-stage finetune classifier. For example, on CIFAR100-LT (IF=100), Our method achieves \textbf{55.88\%} in the first stage, outperforming the two-stage weight rebalancing strategy (53.35\%). After finetuning the classifier with MaxNorm, our method achieves \textbf{57.11\%}, which accuracy increased by \textbf{+3.76\%} compared with the previous SOTA. Compared to contrastive learning families, such as PaCo and BCL, GLMC surpasses previous SOTA by \textbf{+5.11\%}, \textbf{+5.73\%}, and \textbf{+7.46\%} under imbalance factors of 100, 50, and 10, respectively. In addition, GLMC does not need a large batch size and long training epoch to pretrain the feature extractor, which reduces training skills.

\noindent\textbf{Results on ImageNet-LT.}  Table \ref{tabel-imageNetLT} compares GLMC with state-of-the-art baselines on ImageNet-LT dataset. We report the Top-1 accuracy on Many-shot, Medium-shot, and Few-shot groups. As shown in the table, with only one-stage training, GLMC significantly improves the performance of the head class by 70.1\%, and the overall performance reaches 56.3\%, similar to PaCo (180 epochs). After finetuning the classifier, the tail class of GLMC can reach 45.5\% (+ MaxNorm \cite{alshammari2022long}) and 42.19\% (+ BS \cite{ren2020balanced}), which significantly improves the performance of the tail class.
\begin{table}
	\centering
	\caption{Top-1 accuracy (\%) on full CIFAR-10 and CIFAR-100 dataset with ResNet-50 backbone.}
	\begin{tabular}{c|c|c} 
		\hline
		Method  & CIFAR-10 & CIFAR-100  \\ 
		\hline
		vanilla & 94.85   & 75.28     \\
		MixUp \cite{zhang2017mixup}  & 95.95   & 77.99     \\
		CutMix \cite{yun2019cutmix} & 95.41   & 78.03     \\
		SupCon \cite{khosla2020supervised} & 96      & 76.5      \\
		PaCo \cite{cui2021parametric}   & -       & 79.1      \\ 
		\hline
		ours    & \textbf{97.23}   & \textbf{83.05}     \\
		\hline
	\end{tabular}
	
	\label{table-full-cifar}
\end{table}
\subsection{Full ImageNet and CIFAR Recognition}
GLMC utilizes a global and local mixture consistency loss as an auxiliary loss in supervised loss to improve the robustness of the model, which can be added to the model as a plug-and-play component. To verify the effectiveness of GLMC under a balanced setting, we conduct experiments on full ImageNet and full CIFAR. They are indicative to compare GLMC with the related state-of-the-art methods (MixUp \cite{zhang2017mixup}, CutMix \cite{yun2019cutmix}, PaCo \cite{cui2021parametric}, and SupCon \cite{cui2021parametric}). Note that under full ImageNet and CIFAR, we remove the cumulative reweighting and resampling strategies customized for long-tail tasks.

\noindent\textbf{Results on Full CIFAR-10 and CIFAR-100.} For CIFAR-10 and CIFAR-100 implementation, following PaCo and SupCon, we use ResNet-50 as the backbone. As shown in Table \ref{table-full-cifar}, on CIFAR-100, GLMC achieves \textbf{83.05\%} Top-1 accuracy, which outperforms PaCo by \textbf{3.95\%}. Furthermore, GLMC exceeds the vanilla cross entropy method by \textbf{2.13\%} and \textbf{7.77\%} on CIFAR-10 and CIFAR-100, respectively, which can significantly improve the performance of the base model.

\noindent\textbf{Results on Full ImageNet.} In the implementation, we transfer hyperparameters of GLMC on ImageNet-LT to full ImageNet without modification. The experimental results are summarized in Table \ref{table-full-img}. Our model achieves \textbf{80.2\%} Top-1 accuracy, outperforming PaCo and SupCon by \textbf{+0.9\%} and \textbf{+1.8\%}, respectively. Compared to the positive/negative contrastive model (PaCo and SupCon). GLMC does not need to construct negative samples, which can effectively reduce memory usage during training.

\subsection{Ablation Study} 
To further analyze the proposed GLMC, we perform several ablation studies to evaluate the contribution of each component more comprehensively. All experiments are performed on CIFAR-100 with an imbalance factor of 100.

\begin{figure}[]
	\centering
	\includegraphics[width=0.45\textwidth]{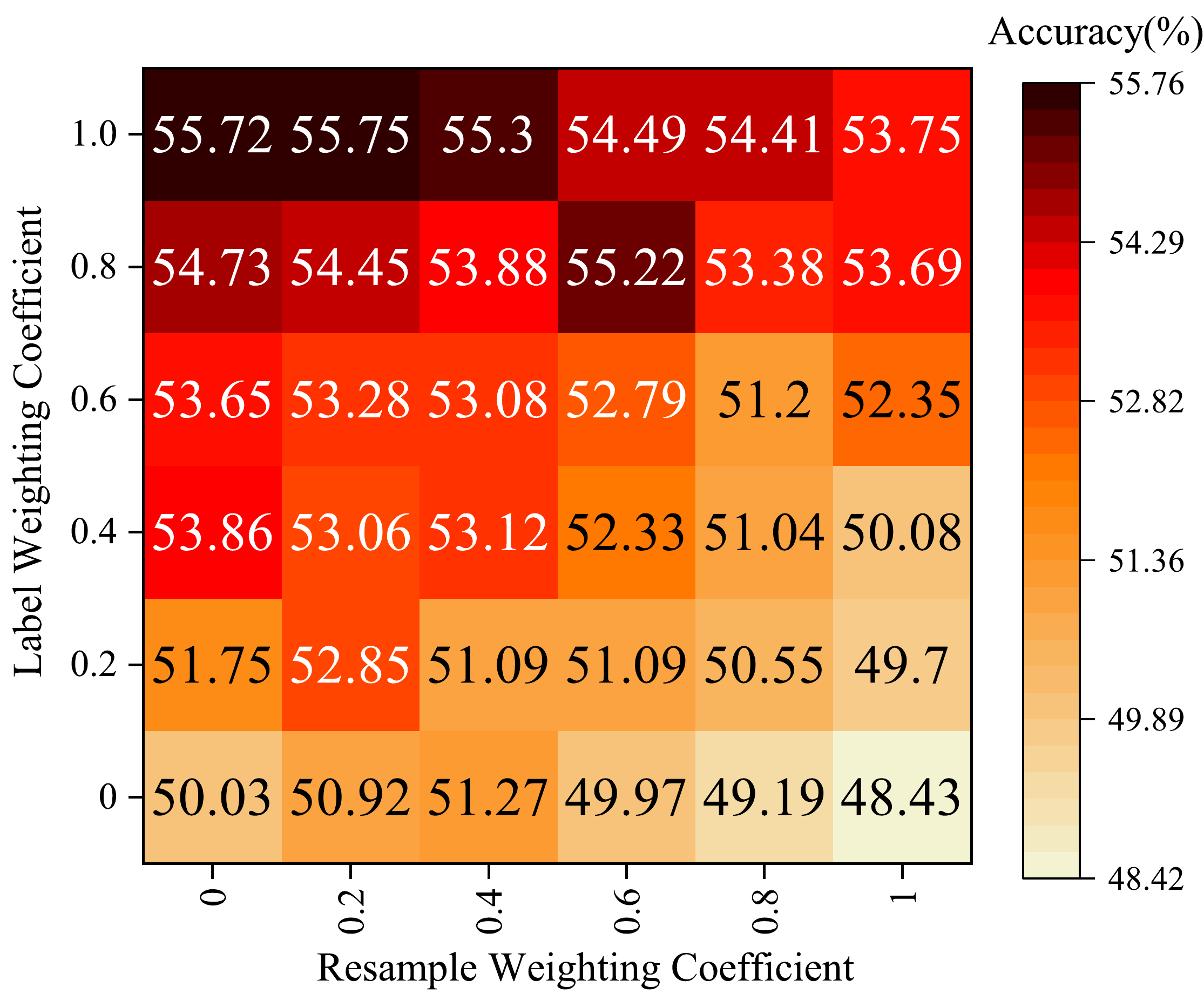}
	\caption{Confusion matrices of different label reweighting and resample  coefficient $k$ on CIFAR-100-LT with an imbalance ratio of 100.}
	\label{fig:ablation-label}    
\end{figure}

\noindent \textbf{The effect of rebalancing intensity.} As analyzed in Sec. \ref{class-balanced learning}, we mitigate head classes bias problems by reweighting labels and sampling weights by inverting the class sampling frequency. See Fig.\ref{fig:ablation-label}, we set different reweighting and resampling coefficients to explore the influence of the rebalancing strategy of GLMC on long tail recognition. One can see very characteristic patterns: the best results are clustered in the upper left, while the worst are in the lower right. It indicates that the class resampling weight is a very sensitive hyperparameter in the first-stage training. Large resampling weight may lead to model performance degradation, so it should be set to less than 0.4 in general. And label reweighting improves long tail recognition significantly and can be set to 1.0 by default.

\begin{figure}[]
	\centering
	\includegraphics[width=0.45\textwidth]{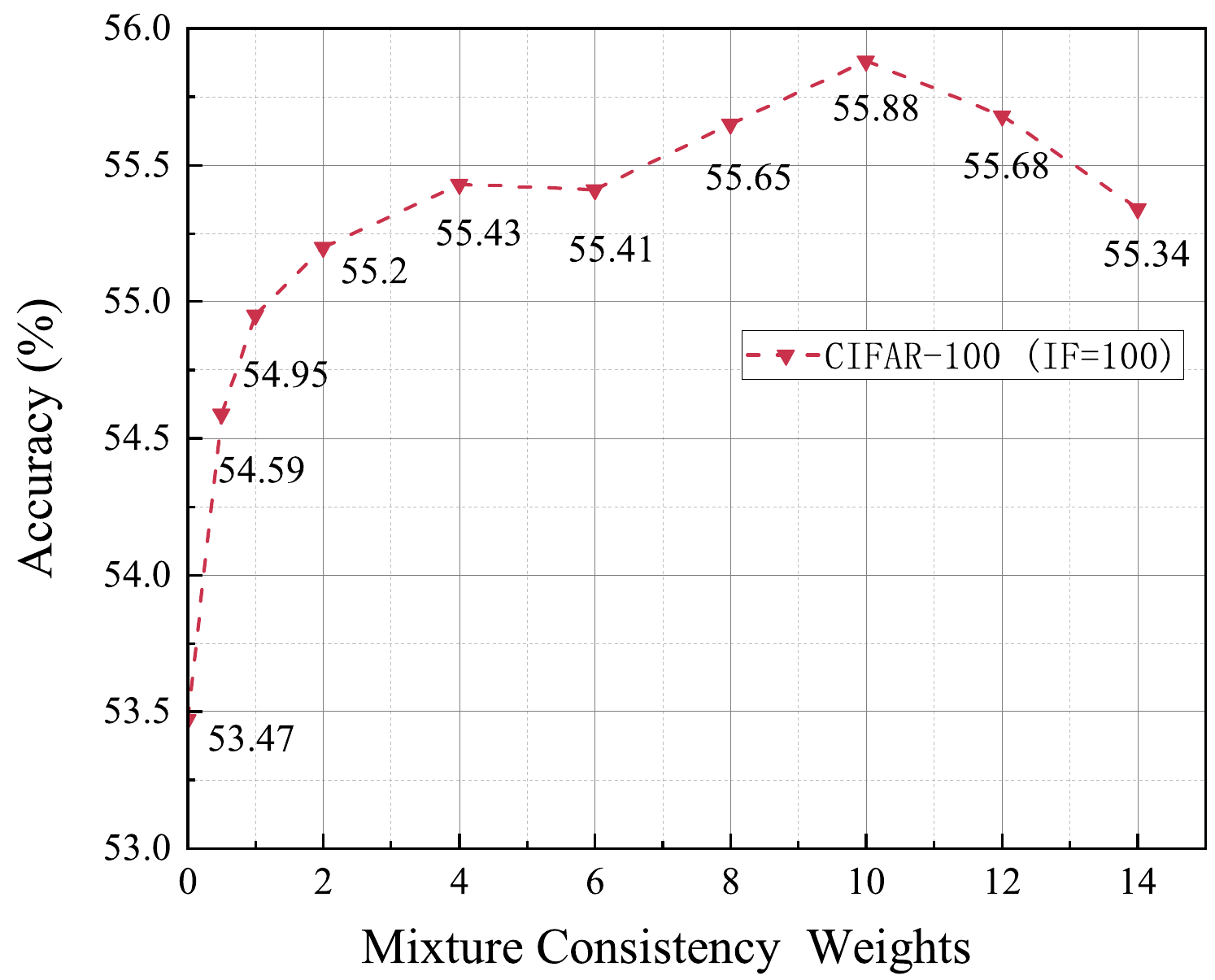}
	\caption{Different global and local mixture consistency weights on CIFAR-100-LT (IF = 100) .}
	\label{fig:ablation-contrast_w}    
\end{figure}

\noindent \textbf{The effect of mixture consistency weight $\gamma$.} We investigate the influence of the mixture consistency weight $\gamma$ on the CIFAR100-LT (IF=100) and plot the accuracy-weight curve in Fig.\ref{fig:ablation-contrast_w}. It is evident that adjusting $\gamma$ is able to achieve significant performance improvement. Compared with the  without mixture consistency ($\gamma$=0), the best setting ($\gamma$=10) can improve the performance by +2.41\%.

\begin{table}[] 
	
	\begin{center}
		\caption{Ablations of the different key components of GLMC architecture. We report the accuracies (\%) on CIFAR100-LT (IF=100) with ResNet-32 backbone. Note that all model use one-stage training. } 
		\begin{tabular}{c|c|c}
			\hline
			\begin{tabular}[c]{@{}c@{}}Global and Local \\Mixture Consistency\end{tabular}  &\begin{tabular}[c]{@{}c@{}} Cumulative \\Class-Balanced \end{tabular}  & Accuracies(\%) \\ \hline
			$\times$   & $\times$   &   38.3 \\ 
			
			$\times$    & \checkmark   &  44.63   \\ 
			\checkmark   & $\times$   &   50.11  \\ 
			\checkmark   & \checkmark  & 55.88    \\ \hline

		\end{tabular}

		\label{ablation_component}
	\end{center}
\end{table}

\noindent \textbf{The effect of each component.} GLMC contains two essential components:(1) Global and Local Mixture Consistency Learning and (2) Cumulative Class-Balanced reweighting. Table \ref{ablation_component} summarizes the ablation results of GLMC on CIFAR100-LT with an imbalance factor of 100. Note that both settings are crossed to indicate using a standard cross-entropy training model. We can see that both components significantly improve the baseline method. Analyzing each element individually, Global and Local Mixture Consistency Learning is crucial, which improves performance by an average of 11.81\% (38.3\% $\rightarrow$ 50.11\% ).

\section{Conclusion}
In this paper, we have proposed a simple learning paradigm called Global and Local Mixture Consistency cumulative learning (GLMC). It contains a global and local mixture consistency loss to improve the robustness of the feature extractor, and a cumulative head-tail soft label reweighted loss mitigates the head class bias problem. Extensive experiments show that our approach can significantly improve performance on balanced and long-tailed visual recognition tasks.

\noindent \textbf{Acknowledgements.} The authors would like to acknowledge the financial support provided by the Natural Science Foundation of China (NSFC) under Grant: No.61876166, Yunnan Basic Research Program for Distinguished Young Youths Project under Grant: 202101AV070003, and Yunnan Basic Research Program under Grant: 202201AS070131.

{\small
\bibliographystyle{ieee_fullname}
\bibliography{egbib}
}

\end{document}